\documentclass[10pt,twocolumn,letterpaper]{article}
\usepackage[arxiv]{wacv}

\usepackage{times}
\usepackage{epsfig}
\usepackage{graphicx}
\usepackage{amsmath}
\usepackage{amssymb}
\usepackage{subcaption}
\usepackage{booktabs}
\usepackage{multirow}
\usepackage[export]{adjustbox}

\usepackage[pagebackref=true,breaklinks=true,letterpaper=true,colorlinks,bookmarks=false]{hyperref}
\usepackage[capitalize]{cleveref}
\crefname{section}{Sec.}{Secs.}
\Crefname{section}{Section}{Sections}
\Crefname{table}{Table}{Tables}
\crefname{table}{Tab.}{Tabs.}
\Crefname{equation}{Equation}{Equations}
\crefname{equation}{Eq.}{Eqs.}

\def\wacvPaperID{1377} 
\def\confName{WACV}
\def\confYear{2024}

\begin{document}

\title{Flexible Techniques for Differentiable Rendering with 3D Gaussians}

\author{Leonid Keselman \qquad Martial Hebert\\
Carnegie Mellon University\\
Pittsburgh, PA, USA\\
{\tt\small \{lkeselma,hebert\}@cs.cmu.edu}
}

\maketitle

\begin{abstract}
   Fast, reliable shape reconstruction is an essential ingredient in many computer vision applications.  Neural Radiance Fields demonstrated that photorealistic novel view synthesis is within reach, but was gated by performance requirements for fast reconstruction of real scenes and objects. Several  recent approaches have built on alternative shape representations, in particular, 3D Gaussians. We develop extensions to these renderers, such as integrating differentiable optical flow, exporting watertight meshes and rendering per-ray normals. Additionally, we show how two of the recent methods are interoperable with each other. These reconstructions are quick, robust, and easily performed on GPU or CPU. For code and visual examples, see \url{https://leonidk.github.io/fmb-plus}. 
\end{abstract}

\section{Introduction}\label{sec:intro}
As computer vision systems are more widely deployed in society, either on robots or via mixed reality headsets, users will desire that reconstructions of their many regular everyday objects. While classic techniques from multiview scene reconstruction could be used~\cite{schoenberger2016mvs}, modern approaches strive for more photorealistic scene generation, such as those created by Neural Radiance Fields (NeRF)~\cite{nerf20} and the large array of follow-up work~\cite{Tancik_2023}. Of note, these NeRF approaches could be seen as differentiable renderers~\cite{liu2019soft,DSS_points}, where an underlying scene representation is optimized for view synthesis. However, NeRF methods are  demanding in their computational requirements, even with speed-up methods such as Instant-NGP~\cite{mueller2022instant}. 

\begin{figure}[thp]
    \centering
    \begin{subfigure}[t]{\linewidth}
        \centering
        \includegraphics[width=0.85\linewidth]{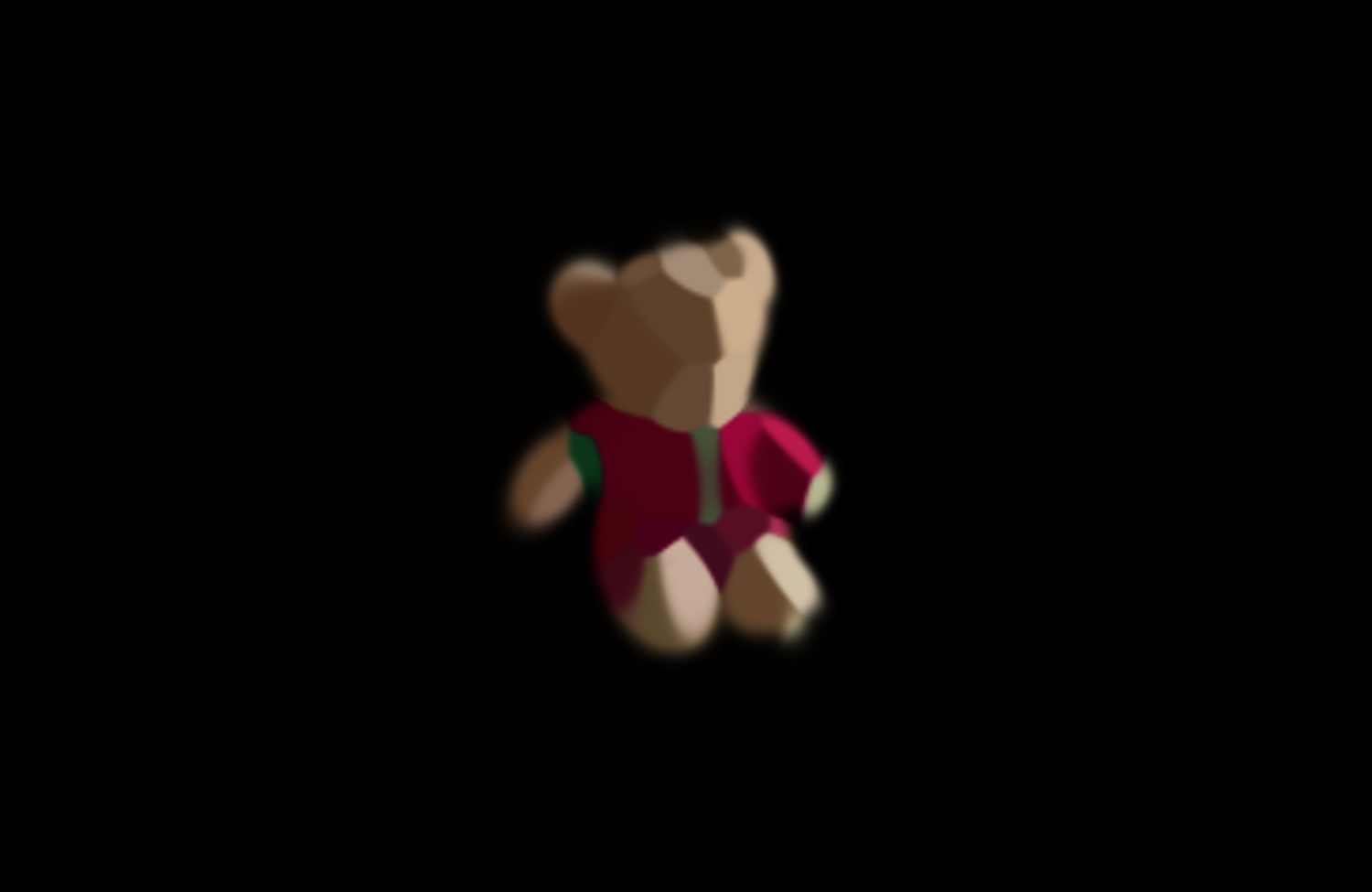}
        \caption{\textbf{Fuzzy Metaballs}~\cite{keselman2022fuzzy} raytraced dozens of Gaussians, used random initialization, and estimated geometry and pose for objects.}
    \end{subfigure}
    \begin{subfigure}[t]{\linewidth}
        \centering
        \includegraphics[width=0.85\linewidth]{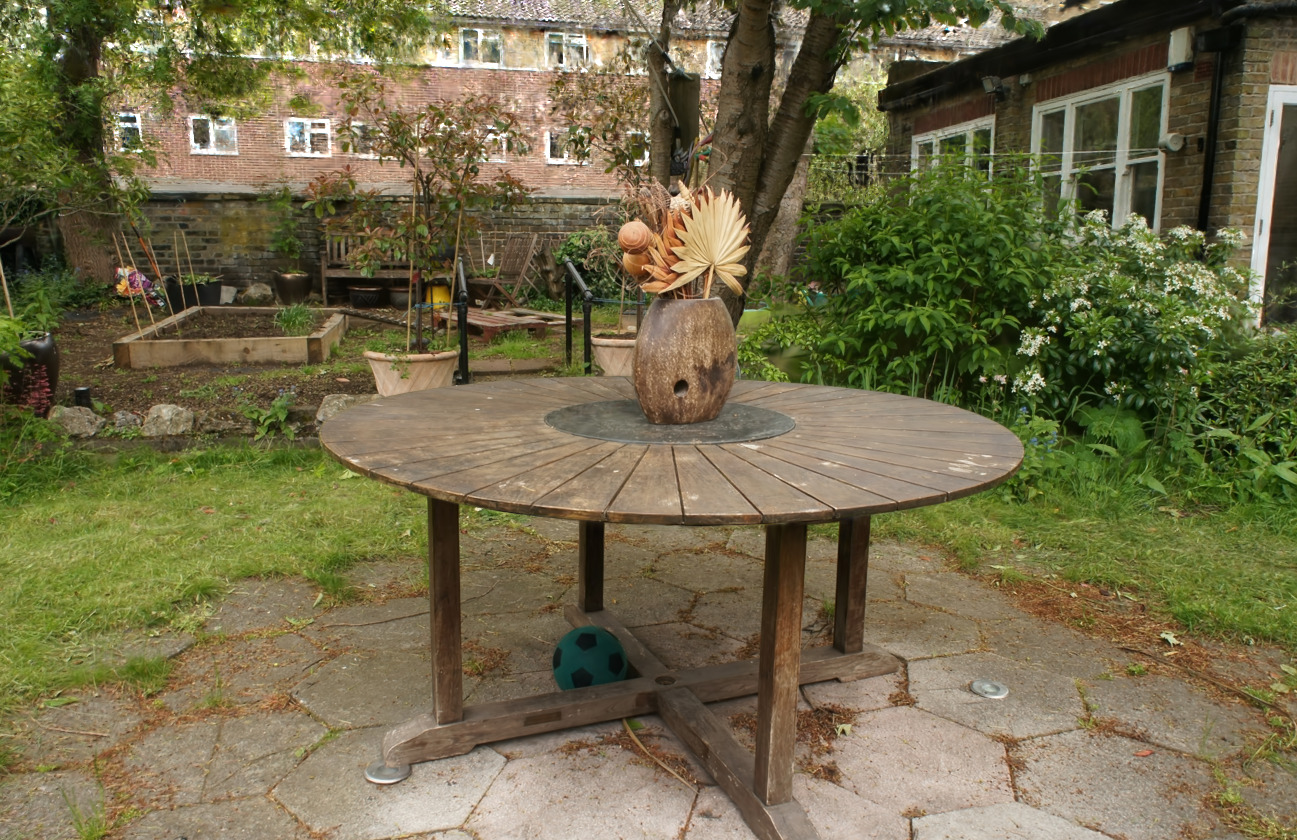}
        \caption{\textbf{3D Gaussian Splatting}~\cite{Kerbl2023splat} splatted millions of Gaussians, used SfM~\cite{schoenberger2016mvs} initialization, and synthesized novel views for scenes.}
    \end{subfigure}
\caption{\textbf{Recent approaches in differentiable rendering use 3D Gaussians as an underlying representation.}. These approaches enable fast reconstruction of 3D objects and scenes. This paper demonstrates how both approaches use the same underlying representation and how to enhance and export such reconstructions.  }
\label{fig:fig1}
\end{figure}

Two recent papers explored similar approaches to fast NeRF alternatives -- Fuzzy Metaballs~\cite{keselman2022fuzzy} in ECCV 2022 and 3D Gaussian Splatting in SIGGRAPH 2023~\cite{Kerbl2023splat}. Both approaches represent geometry using a set of classic primitives, specifically 3D Gaussians~\cite{Heckbert85funwith,10.2307/90667}.  The former built a differentiable raytracer for 3D Gaussians, connected it to the metaball literature~\cite{10.1145/357306.357310}, developed a sort-free rendering function, focused on fast CPU runtimes, used dozens of Gaussians, and performed quantitative experiments for object reconstruction and  pose estimation. A year layer, the latter designed a fast rasterizer with a custom CUDA kernel for splatting~\cite{metaballCompositing,shadowSplatting01}, used millions of Gaussians, and focused their system on reconstructing entire scenes to approach NeRF levels of fidelity. We demonstrate connections between and extensions to both methods. 

This paper is focused on extending the Fuzzy Metaballs renderer to make it simpler, more robust, and to add additional features. We show that these recent papers are interoperable and render the same underlying representation. Since most well-established rendering techniques are built on triangle meshes, we demonstrate a reliable way to transform 3D Gaussian representations into meshes.

We summarize our contributions as follows:
\begin{itemize}
    \item Develop a simplified version of Fuzzy Metaballs~\cite{keselman2022fuzzy} for shape reconstruction from Gaussians (\Cref{sec:twoparam}).
    \item Show how Fuzzy Metaballs~\cite{keselman2022fuzzy} can be rendered without hyperparameters  (\Cref{sec:alphac}).
    \item Show how to get per-pixel, differentiable optical flow and its benefits in reconstruction (\Cref{sec:flow}).
    \item Demonstrate how to export meshes from shapes defined by  3D Gaussians (\Cref{sec:mesh}).
    \item Show that existing 3D Gaussian rendering methods~\cite{Kerbl2023splat,keselman2022fuzzy} are interoperable (\Cref{sec:interop}).
    \item Develop a loss-based approach to reparameterizing Gaussians by splitting components (\Cref{sec:splits}).
\end{itemize}

These techniques enable more powerful, flexible uses of these 3D Gaussian shape representations for applications involving shape reconstruction. Of note, these approaches require no pretraining and can be optimized directly on the scene of interest without dataset bias, making them applicable to robotics applications where robots may be interacting with novel objects, and are limited by onboard compute. 


\section{Related Work}\label{sec:related}
A comprehensive overview of related work can be found in prior papers using 3D Gaussians, including Fuzzy Metaballs~\cite{keselman2022fuzzy}, 3D Gaussian Splatting~\cite{Kerbl2023splat} and VoGE~\cite{wang2023voge}.

Some early methods of building models from partial observations used generalized cylinders~\cite{10.5555/905981}. More commonly, methods build on top of triangle meshes, point clouds and surfels~\cite{10.1145/344779.344936}. Differentiable renders have been built for these representations, initially for meshes~\cite{kato2017neural,liu2019soft,OpenDR}. These include custom backends that allow for fast GPU-based results~\cite{Laine2020diffrast}, and high-quality results~\cite{Mitsuba2019diff}. Other works focus on point clouds~\cite{insafutdinov2018unsupervised,DSS_points}. Pulsar~\cite{lassner2020pulsar} uses spheres as its representation, which are equivalent to isotropic 3D Gaussians. Primitives-based rendering methods have benefits for both for composition~\cite{zhang2022nerfusion} and tracking~\cite{luiten2023dynamic}.

The earliest work using 3D Gaussians in rendering came from Blinn~\cite{10.1145/357306.357310}, originally called atoms, blobs or metaballs, and were the birth of implicit surfaces. Several methods these techniques~\cite{gourmel:hal-01516266,horvath-2018-ism,10.1145/122718.122743,Szcsi2012RealTimeMR,anisotropicGaussianRender1990,Wyvill1986,raytraceSoft}. Some renderers used rays while others used splatting~\cite{ALD2006PSIROPSD}, and some recent differentiable renders build screen space Gaussians~\cite{sampling2022ECCV}. Others use Gaussians as a primary representation in computer vision
~\cite{Eckart2018,Eckart2015,Eckart2016,genova2020local,10068771,hertz2019pointgmm,Magnusson_2009} or render them via projection, search, or other techniques~\cite{9126150,8586902,shankar20mrfmap}. 

Gaussians can be seen as a fundamental building block that only uses the 1st ($\mu$) and 2nd ($\Sigma$) order moments~\cite{847726}. Point clouds only use $\mu$. Oriented point clouds add a covariance  eigenvector~\cite{Eckart2018}, and Gaussian Mixtures~\cite{Eckart2016} use all the information. Connections in this space include error metrics~\cite{Garland1997,Mahalanobis1936OnTG} and physics simulations~\cite{fastCovariance}. 

There is work on connecting NeRF-style differentiable renderers to mesh representations. MobileNeRF~\cite{chen2022mobilenerf} used the rasterization pipeline of commodity hardware to perform rendering of NeRF-like objects. VMesh~\cite{guo2023vmesh} constructed a hybrid representation of volume and mesh. NeRFMeshing~\cite{rakotosaona2023nerfmeshing} learned a Signed Surface Approximation Network to distill NeRF representations into meshes. 

\section{Ray-Shape Intersections}\label{sec:rayshape}
Existing methods for intersecting 3D Gaussians and rays typically take two forms. Methods taking inspiration from NeRF family methods~\cite{Kerbl2023splat,wang2023voge} typically sort all intersections and use closer Gaussians to attenuate the contributions of further Gaussians. Fuzzy Metaballs~\cite{keselman2022fuzzy} introduced a heuristic technique for blending intersections which did does not require sorting. \Cref{sec:weightb} summarizes this technique, \Cref{sec:twoparam} presents a simplification, and \Cref{sec:alphac} proposes a variation without hyperparameters. All three correctly render the same shapes, so objects optimized by one can be reproduced with the other.


\subsection{Weighted Blending}\label{sec:weightb}
In Fuzzy Metaballs~\cite{keselman2022fuzzy}, intersections between each ray ($\vec{v}$) and each Gaussian are computed separately and then combined with a weighted average. Each Gaussian is parameterized as mean ($\mu \in \mathbb{R}^{3}$), inverse root precision ($\Sigma^{-\frac{1}{2}}$) and a weight ($\lambda_i \geq 0$) where the $\sum_i \lambda_i = 1$. 

Multivariate Gaussians are defined as 
\begin{equation}\label{eq:component}
    P(\vec{x}) = |\Sigma|^{-\frac{1}{2}} \exp\left(-\frac{1}{2} (\vec{x}-\mu)^T \Sigma^{-1} (\vec{x}-\mu)\right),
\end{equation}
and the unnormalized log distance for each Gaussian is
\begin{equation}\label{eq:quad_form}
    d(\vec{v} t) = - \frac{1}{2} \left[ (v t -\mu)^T \Sigma^{-1} (v t-\mu)\right] + \log(\lambda_i),
\end{equation}
which we will refer to as $d_i$ when referring to evaluating the $i$-th Gaussian for a ray $\vec{v}$ (here we use the point of maximum likelihood, $ t_i = \frac{\mu_i^T \Sigma_i^{-1} v}{v^T \Sigma_i^{-1} v}$, the linear approach~\cite{keselman2022fuzzy}). 

For each Gaussian, its intersection ($t_i$) and distance ($d_i$) are used to obtain weights ($w_i$). The final intersection is obtained as $t_f$, similar to OIT~\cite{SA09,McGuire2013Transparency}:
\begin{equation}\label{eq:blend1}
    t_f = \frac{1}{\sum_i w_i} \sum_i w_i t_i.
\end{equation}

Per-ray estimation of other properties can continue to use \Cref{eq:blend1}. For example, $t_i$ (distance from camera) can be replaced with $\vec{n_i}$ (normal) or $\vec{c_i}$ (color) and the same blending functions can be reused. For more details about normal computation, see ~\Cref{sec:mesh}. It is also helpful to think of the unnormalized Gaussian density as:

\begin{equation}\label{eq:delta}
\delta_i = \exp(d_i),
\end{equation}
The original Fuzzy Metaballs approach uses 5 hyperparameters ($\beta_1,\beta_2,\beta_3,\beta_4,\beta_5$) to compute the weights and the transparency. There is also a shape scale ($\eta$) to account for shapes of different scales and return identical results. The following are the weight and $\alpha$ (quality/opacity)  functions:

\begin{equation}\label{eq:weight_old}
    w_i = \exp\left(\beta_1 d_i \sigma\left(\frac{\beta_3}{\eta} t_i\right) -\frac{\beta_2}{\eta} t_i\right),
\end{equation}

\begin{equation}\label{eq:alpha_old}
    \alpha = \sigma\left(\beta_4  \sum_i \delta_i + \beta_5\right).
\end{equation}

\subsection{Two Parameter Model}\label{sec:twoparam}
In shape reconstruction from video, 3 of these hyper-parameters are not necessary and we develop a simplified two parameter model that works in reconstruction settings. 

First, $\beta_3$ was used to give some minor contribution to intersections behind the camera. While this does make the renderer more differentiable, it is not used in practice, so the $\sigma$ function can be removed, leaving us with

\begin{equation}\label{eq:weight_new}
    w_i = \exp\left(\beta_1 d_i  -\frac{\beta_2}{\eta} t_i\right).
\end{equation}
Since the original paper~\cite{keselman2022fuzzy} focused on rendering full, proper Gaussian Mixture Models (where $\sum_i \lambda_i =1$), it required a normalizing factor $\beta_4$ to account for how much opaque a GMM should be. In the case of reconstruction, there is no need for strict GMMs, so learning $\beta_4$ can be the responsibilities of unnormalized $\lambda_i$. Additionally, the GMM focus suggested a smooth step function (sigmoid) for computing $\alpha$, but shape reconstruction can simplify this to a decaying exponential, letting us drop the intercept term $\beta_5$. This allows us to have a simple, perhaps familiar~\cite{nerf20,10.1145/3450626.3459815}, expression for $\alpha$:
\begin{equation}\label{eq:alpha_new}
    \alpha =  1- \exp\left(-\sum_i \delta_i \right) .
\end{equation}

We use this simplified two parameter model in most of our experiments. While we noticed some small differences due to losing the global normalization condition, they were minor and mostly in the space of pose estimation, which is not the focus of our experiments here for shape estimation. 

\subsection{Zero Parameter Model}\label{sec:alphac}
With the simplified design given above, and the NeRF-style approaches in other 3D Gaussian papers~\cite{Kerbl2023splat,wang2023voge}, we also investigate an alpha compositing variant of Fuzzy Metaball rendering that requires sorting all intersections and then computing transmission. The transmissions can be seen as weights and \Cref{eq:blend1} can be used to compute depth estimates (as well as normal and color estimates):

\begin{equation}\label{eq:weight_alpha}
    w_i = T_i \left( 1 - \exp(- \delta_i )\right)
\end{equation}
where $T_i$ is the accumulated transmissions of earlier intersections
\begin{equation}\label{eq:transmission}
    T_i = \exp\left(- \sum_j \delta_j 1[t_j < t_i] \right)
\end{equation}

These equations create a version of differentiable rendering with metaballs that is hyperparameter-free, doesn't require splatting, and enables the same usage of weights for depth, normal and color computation as prior work~\cite{keselman2022fuzzy}. As our experiments in \Cref{tab:runtime_bck} and \Cref{fig:fm_opt} show, this approach is slower but performs nearly identically.

 \begin{figure*}[thp]
    \centering
    \includegraphics[width=0.32\linewidth,frame]{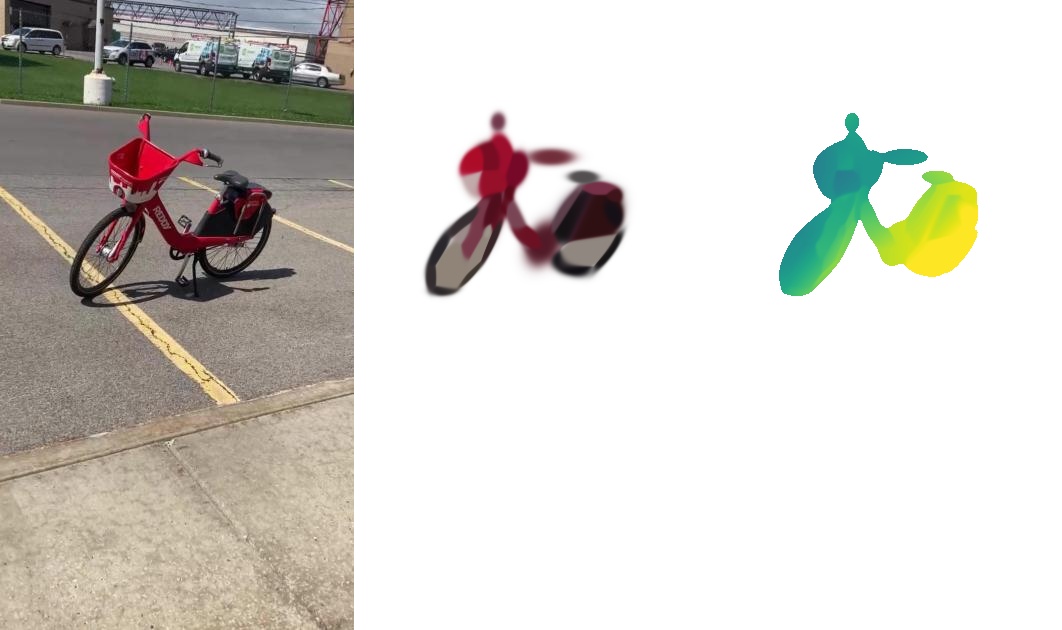}
    \includegraphics[width=0.32\linewidth,frame]{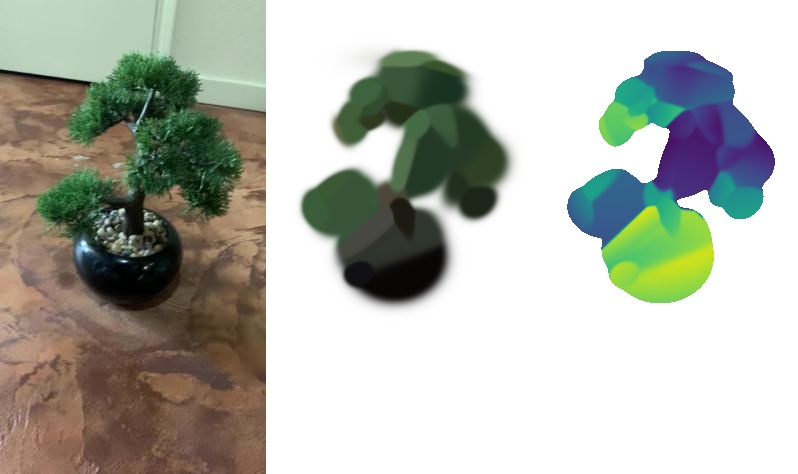}
    \includegraphics[width=0.32\linewidth,frame]{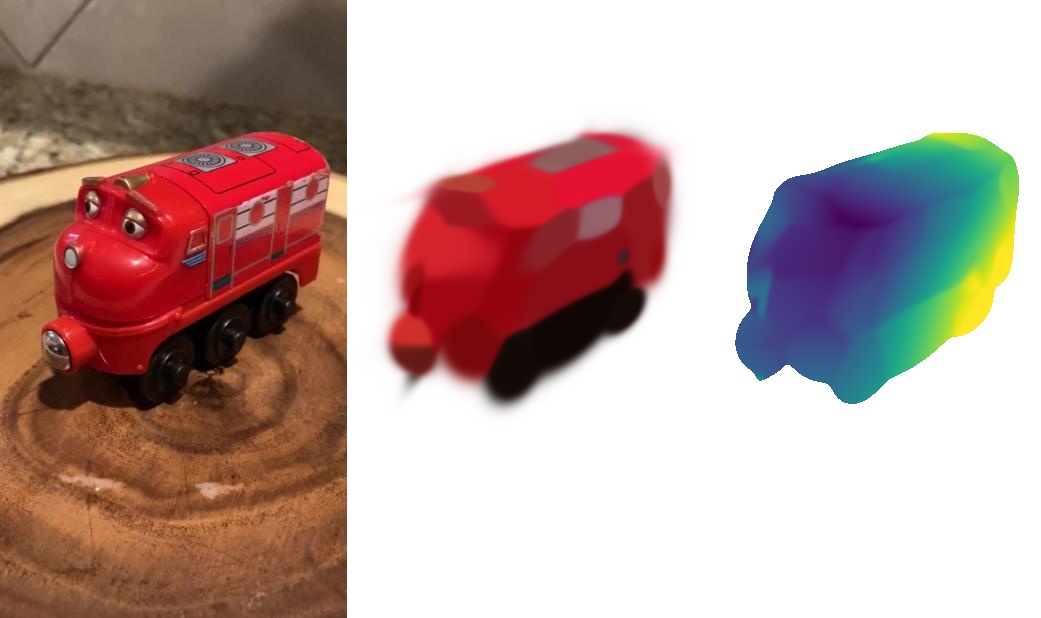}
    \includegraphics[width=0.48\linewidth,frame]{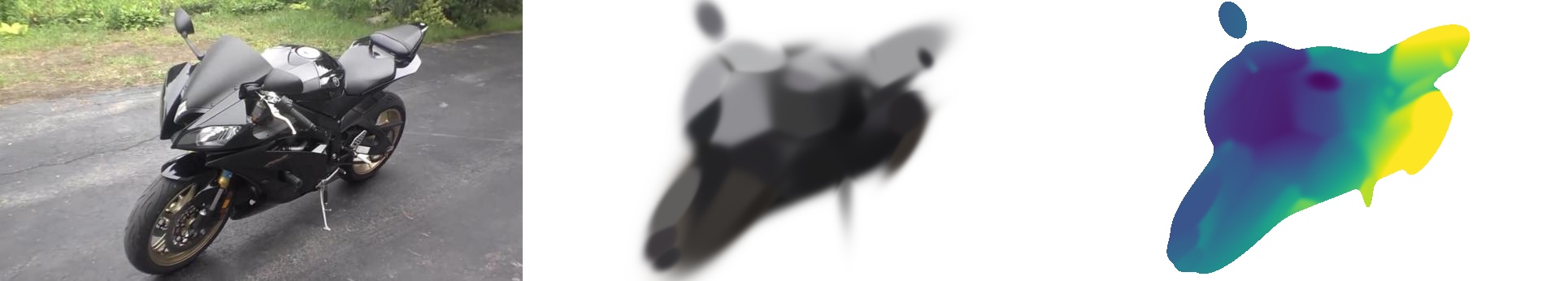}
    \includegraphics[width=0.48\linewidth,frame]{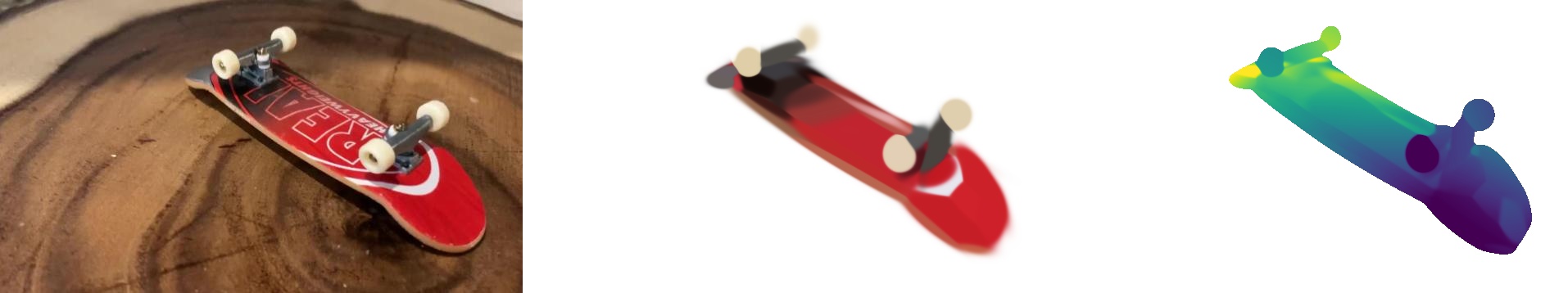}

\caption{\textbf{Shape reconstruction} using 40 3D Gaussians and converging in under one minute, with color. (See \Cref{sec:method} for details). All objects are reconstructed from videos in the CO3D~\cite{reizenstein21co3d} dataset.  }
\label{fig:viz_results}
\end{figure*}

\section{Shape Reconstruction}\label{sec:method}
 The input to the system is a video and a single masked frame. COLMAP~\cite{schoenberger2016mvs} produces poses, XMem~\cite{cheng2022xmem} propagates masks, Unimatch~\cite{xu2023unifying} produces flow, and 3D Gaussians optimized to fit the data. The shape converges quickly since we use a ray-based differentiable renderer and are able to sample minibatches that includes pixels across all frames.

Our differentiable rendering code is based on Fuzzy Metaballs~\cite{keselman2022fuzzy}, which is built in JAX~\cite{jax2018github} and allows for reconstructions on both the CPU and GPU. With an Nvidia GTX 1080, we can do memory for image sequences of roughly $250 \times 130$, while our CPU experiments are typically closer to $125 \times 65$. Both sets of experiments typically converge in less than a minute on commodity hardware. We use 40  Gaussians as in prior work, for ease of comparison.  

Differentiable renders provide flexibility for many loss functions in shape reconstruction. For our experiments, we combine of cross-entropy loss ($L_M$) for objects segmentation masks, and $L_1$ losses for color and flow (\Cref{sec:flow}), weighted by the object mask. Estimated alpha is clipped to $[10^{-6},1-10^{-6}]$. For $\alpha$, color ($c$), and optical flow ($f$) we use the following loss function:
\begin{align}
L_M &= \alpha \cdot \log(\hat{\alpha}) +  (1-\alpha) \cdot \log(1-\hat{\alpha})   \\
L_C &= \alpha \cdot || c - \hat{c} ||_1 \\ 
L_F &= \alpha \cdot || f - \hat{f} ||_1 \\   
L &= L_M + \lambda_C L_C + \lambda_F L_F \label{eq:loss}
\end{align}
In practice, we use colors that have had gamma corrected back to linear intensity, and use a sigmoid to map from an unconstrained parameterization to $[0,1]$. Flow is measured in the scale of half the shorter image dimension. This lets us set $\lambda_C = 4.5$ and $\lambda_F = 210$ for all of our experiments. We used identical settings for all our experiments: initialization is from a randomized small sphere of Gaussians, and we use fixed parameters for learning rate, canonical rescaling, batch size, and all other known parameters. Learning rates are automatically decayed based on statistical criteria~\cite{keselman2022fuzzy}. 

We use the Adam~\cite{kingma2014Adam} optimizer and rescale all scenes to a canonical size based on camera pose distances to balance the optimization of means and precisions~\cite{keselman2022fuzzy} that occurs when using re-scaling optimizers. We randomly sample minibatches of 50,000 rays from across the entire sequence, which leads to extreme fast convergence, often getting a reasonable shape in well under an epoch (\Cref{fig:fm_opt}). We use the per-Gaussian simple colors used in prior work~\cite{keselman2022fuzzy} for simplicity, but extensions to Spherical Harmonics are possible for greater color fidelity per Gaussian~\cite{yu_and_fridovichkeil2021plenoxels,Kerbl2023splat}. 

Examples of our reconstructions can be seen \Cref{fig:viz_results}. Even though the optimization only takes minutes, operates on reasonably low resolution images, with reasonably few Gaussians, we can still see good results. Depth estimates are able to capture small geometric details (notice the kickstand and both mirrors on the motorcycle). Despite the initialization being a small, invisible sphere of Gaussians, the optimization procedure is able to reconstruct shapes with rich geometry (such as the bicycle and the plant). Lastly, the color results look reasonably realistic. Although separate Gaussians must be used to paint 2D flat textures onto surfaces, reasonable results are obtained for the toy truck and the skateboard reconstructions. In the toy truck, grey side and roof panel details appear in the reconstruction. In the skateboard, the painted curve shape is also approximately modeled in the reconstruction, as are both wheels. 

We obtained similarly good results with both \Cref{sec:weightb} and \Cref{sec:alphac}, but all results in \Cref{fig:viz_results} use the faster, two parameter model for optimization and the resulting visuals.

\section{Reconstructing with Optical Flow}\label{sec:flow}
Many approaches to 3D reconstruction focus only on reconstructing the independent images from the given sequence~\cite{Kerbl2023splat}, including all the 3D Gaussian methods~\cite{Kerbl2023splat,keselman2022fuzzy,wang2023voge}. However, in practice, these image sequences are often collected by cell phone videos~\cite{reizenstein21co3d} and have a strong temporal prior. Inspired by work in reconstructing 4D scenes~\cite{li2020neural,yang2021lasr,yang2021viser}, we leverage optical flow in producing more precise 3D reconstructions of static objects. %

Optical flow provides a hypothesis of surface correspondence, which regularizes the shape reconstruction. Correspondence can be essential in classic techniques for shape estimation~\cite{580394,10.1117/12.965752}. Sparse particle video trackers extend this, and obtain long-term video correspondences~\cite{harley2022particle}. 

Optical flow is a useful signal since it is a local estimator, and is robust to the long-term lighting changes that occur when typical users capture scenes under auto-exposure~\cite{jun2022hdr}. This makes it perhaps a more appropriate loss term than color models, which would be sensitive to  lighting changes. Additionally, color often means texture, which implies high frequency texture details that can be difficult to reconstruct~\cite{nerf20}, and perhaps efficient shape estimation can do without. 

\begin{figure}[thpb]
    \centering
    \begin{subfigure}[t]{0.49\linewidth}
    \includegraphics[width=0.49\linewidth,trim={0 80 0 80},clip]{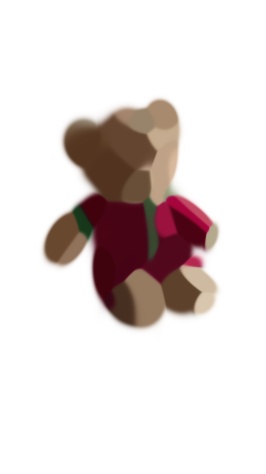}
    \includegraphics[width=0.49\linewidth,trim={0 80 0 80},clip]{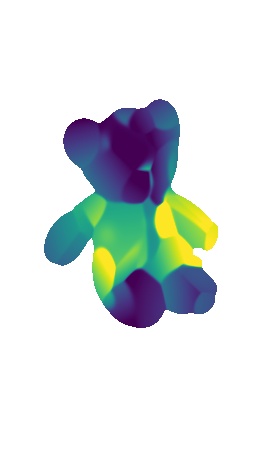}
    \caption{\textbf{Without Flow} }
        \label{fig:without_flow_recon}
    \end{subfigure}
\begin{subfigure}[t]{0.49\linewidth}
    \includegraphics[width=0.49\linewidth,trim={0 80 0 80},clip]{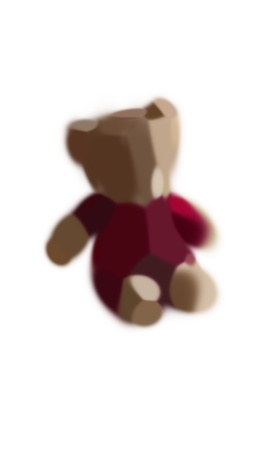}
    \includegraphics[width=0.49\linewidth,trim={0 80 0 80},clip]{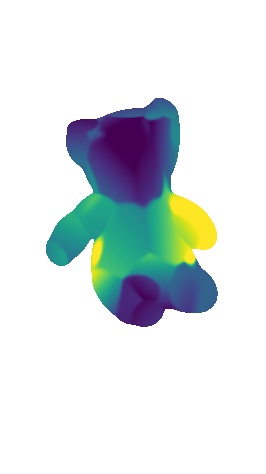}
    \caption{\textbf{With Flow} }
    \label{fig:with_flow_recon}
    \end{subfigure}
\caption{\textbf{Optical flow improves reconstruction}. We show the results of fitting a set of 3D Gaussians to a CO3D~\cite{reizenstein21co3d} sequence. Without a flow loss term, colors are estimated well but the shape is not. After adding flow, color fidelity is sacrificed but shape estimation improves. }
\label{fig:flowout}
\end{figure}

\begin{figure}[thbp]
    \centering
    \begin{subfigure}[t]{0.24\linewidth}
        \centering
        \includegraphics[width=\linewidth,trim={0 80 0 80},clip]{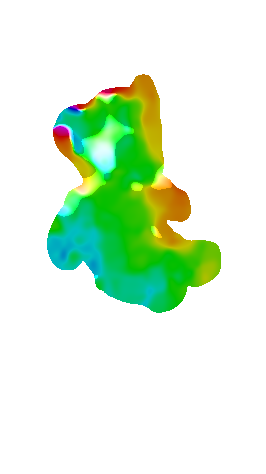}
        \caption{Gunnar~\cite{gunnar2003flow}}
        \label{fig:classic_flow}

    \end{subfigure}
    \begin{subfigure}[t]{0.24\linewidth}
        \centering
        \includegraphics[width=\linewidth,trim={0 80 0 80},clip]{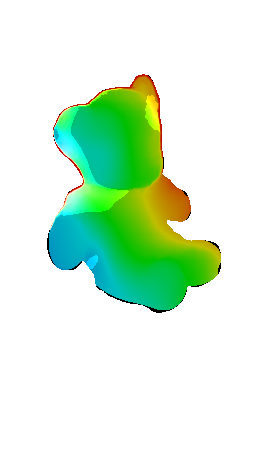}
        \caption{Uni~\cite{xu2023unifying}}
        \label{fig:learned_flow}

    \end{subfigure}
        \begin{subfigure}[t]{0.24\linewidth}
        \centering
        \includegraphics[width=\linewidth,trim={0 80 0 80},clip]{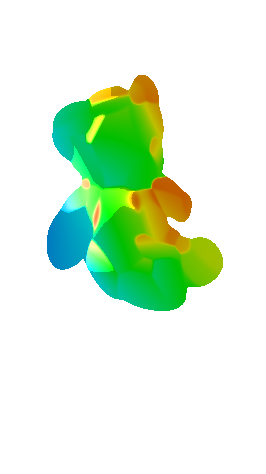}
        \caption{Without}
        \label{fig:without_flow}

    \end{subfigure}
        \begin{subfigure}[t]{0.24\linewidth}
        \centering
        \includegraphics[width=\linewidth,trim={0 80 0 80},clip]{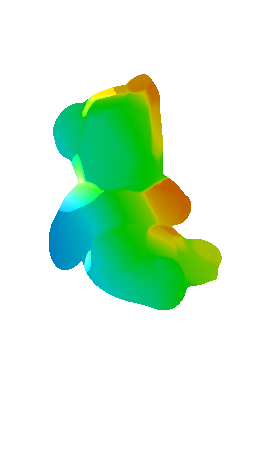}
        \caption{With}
        \label{fig:with_flow}
    \end{subfigure}
\caption{\textbf{Optical Flow Quality}. (a) shows a classic optical flow estimate. (b) shows a modern, learned optical flow estimate. (c) and (d) show the estimated flow for the same frame after fitting 3D Gaussian model; respectively without adding a flow loss and after adding a flow loss term to fit the estimate from (a). Using standard optical flow coloring~\cite{opticalFlowColor2007}.  }
\label{fig:flowcomp}
\end{figure}

The benefits of flow can be from our experiments,  qualitatively in \Cref{fig:flowcomp} and quantitatively in \Cref{tab:flow_quality}. One interesting result is that even classic optical flow~\cite{gunnar2003flow} provides helpful cues to the shape optimization, even though it is very noisy (see~\Cref{fig:classic_flow}). Even better, state-of-the-art flow methods such as Unimatch~\cite{xu2023unifying} are extremely fast (real-time on GPU hardware) and have learned good priors even in texture-less areas. Using such learned flow maps (\Cref{fig:learned_flow}) can help shape estimation greatly. After optimization, predicted model flow (\Cref{fig:with_flow}) is very similar to the given flow estimate from the network that was used to supervise it. Incorporating flow also produces significantly smoother depth maps, as can be seen in \cref{fig:flowout}. 


In our experiments, optical flow slightly hurts the color fidelity of the reconstructed models, but provides much more accurate shape reconstruction, as can be seen in \Cref{fig:flowout}. After adding a flow loss term, this surrogate estimate of surface correspondence helps resolve concavity/convexity ambiguity from the silhouette and color loss terms. For example, the body of the teddy bear becomes smooth and its arms become well defined. 

With the ray-based differentiable rendering of 3D Gaussians, it is reasonably easy to compute per-pixel optical flow. It only requires taking the 3D coordinate obtained from \Cref{eq:blend1}, and transforming it with adjacent camera poses and projecting back into camera coordinates. 

In our implementation, we pass all camera poses into the rendering function and represent the camera with a single inverse focal length parameter (making the assumptions that the images lack distortion, the projection is in the center of the image, and that the pixels are square). After computing the depth image for a given frame, we transform the point cloud (forwards and backwards) and project the transformed coordinates into the image. The changes in coordinates is the direct, per-ray optical flow estimate. Without using another source to regularize the optical flow, we get reasonable estimates but with some artifacts due to shape uncertainty (as can be see in \Cref{fig:without_flow}). 

We include estimates for both forward flow (pose $i$ to pose $i+1$) and backward flow (pose $i$ to pose $i-1$) from our differentiable renderer, for each ray. 

\begin{table}[thpb]
\centering
\begin{tabular}{@{}p{4cm}p{1.4cm}p{1.7cm}@{}}
\toprule
 & \textbf{Depth  Error} &  \textbf{Runtime (seconds)} \\ \midrule
No Color or Flow & 0.271 & 17 \\
Color & 0.262 & 15 \\
Color \& Classic Flow~\cite{gunnar2003flow} & 0.237 & \textbf{14} \\
Color \& Learned Flow~\cite{xu2023unifying} & \textbf{0.155} & 15 \\ \bottomrule
\end{tabular}
\caption{\textbf{Optical flow helps reconstruction} of CO3D sequences~\cite{reizenstein21co3d}. For details see \Cref{sec:flow}.}
\label{tab:flow_quality}
\end{table}

\begin{figure*}[tbhp]
    \centering
\includegraphics[width=0.19\linewidth]{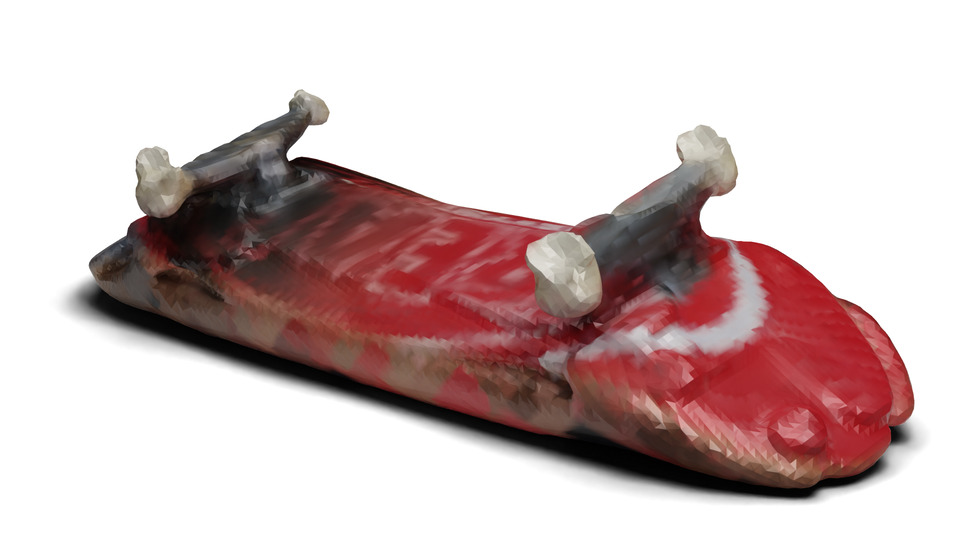} 
\includegraphics[width=0.19\linewidth]{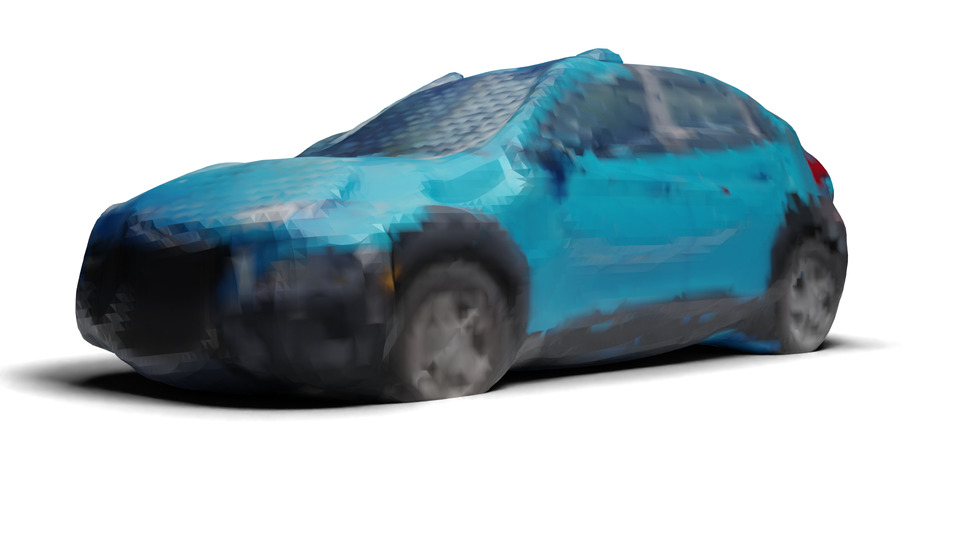} 
\includegraphics[width=0.19\linewidth]{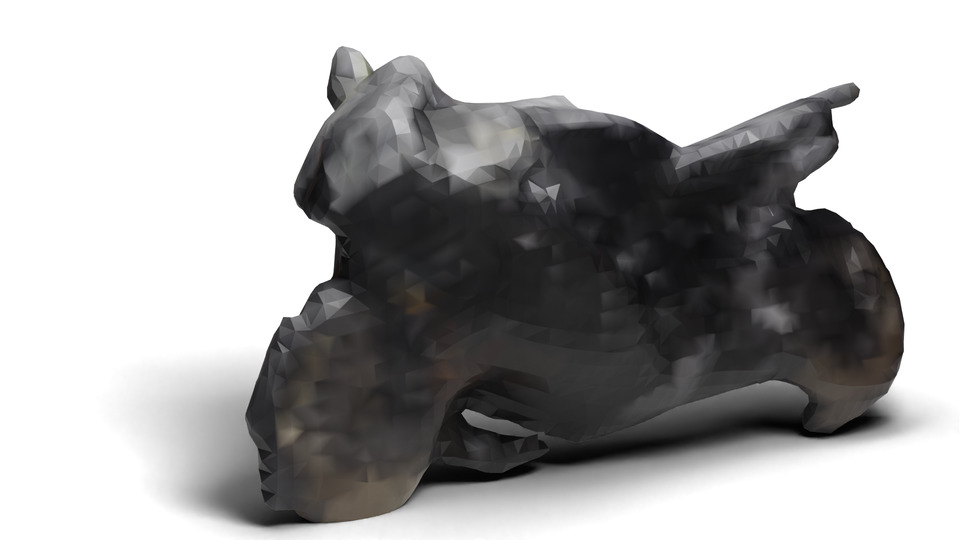} 
\includegraphics[width=0.19\linewidth]{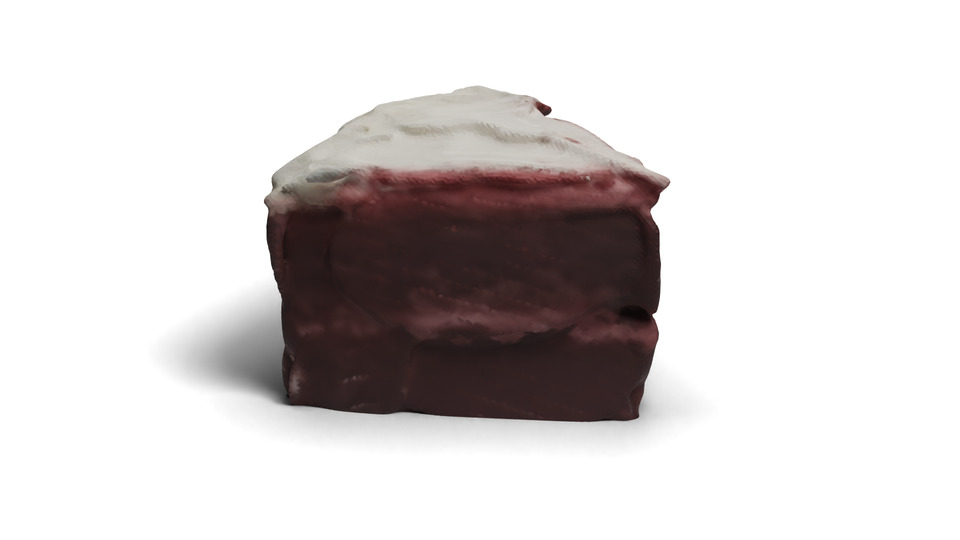}
\includegraphics[width=0.19\linewidth]{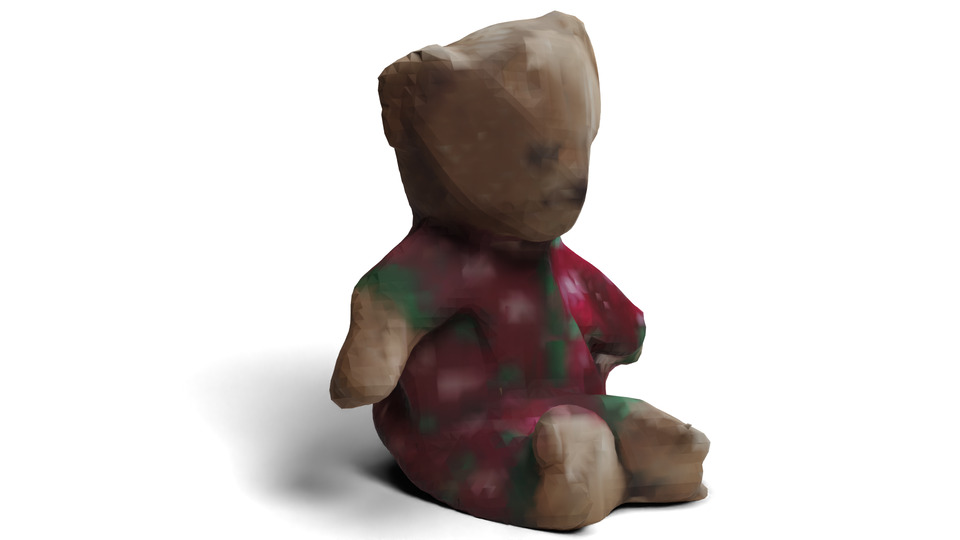} 

\caption{\textbf{Colored Mesh Exports} reconstructed from CO3D sequences~\cite{reizenstein21co3d} with 40 3D Gaussians and exported to Blender (\Cref{sec:mesh}).}
\label{fig:mesh_viz_fm}
\end{figure*}

\section{Exporting Meshes}\label{sec:mesh}
Many successful differentiable renderers realize that fast, efficient, robust differentiable rendering often require smooth, fuzzy and indefinite surface representations~\cite{nerf20,keselman2022fuzzy,Kerbl2023splat}. For optimizing shapes from  videos, it is helpful to have some degree of softness to aid gradient flow. 

On the other hand, commercial rasterization pipelines in most desktop and mobile GPUs typically operate on triangle meshes~\cite{guo2023vmesh,chen2022mobilenerf}. Additional, the field of shape processing often prefers not just definite surfaces in the form of meshes, but watertight meshes~\cite{DBLP:journals/corr/ZhouJ16,DBLP:journals/tog/HuLGCHMM22}. Differentiable mesh renderers typically bridge this divide with an explicit spatial smoothness term~\cite{liu2019soft,ravi2020pytorch3d}. 

Instead, 3D Gaussians, neural surfaces, and other similar methods are implicit surface methods, where the genus of the object can change during optimization. Many existing works export meshes using marching cubes~\cite{lorensen1987marching,nerf20,keselman2022fuzzy}, where a volumetric grid is evaluated and triangles are produced at edges crossing a particular level set threshold. While fast and simple, this can produce non-watertight meshes. Additionally, it can require searching over an ideal threshold to find which level set of the implicit surface matches the explicit surface best~\cite{keselman2022fuzzy}.

In this work, we leverage the surface definition used by the underlying differentiable renderer, and then solve the appropriate Poisson equation~\cite{SGP:SGP06:061-070,kazhdan2013screened}. In Poisson surface reconstruction, oriented point sets (points with normals indicating the local tangent plane of the surface) are used as input to solve a Poisson equation:

\begin{equation} \label{eq:poisson}
    \nabla \cdot \nabla \chi = \nabla \cdot \vec{V} .
\end{equation}
Solving these equations can be done via well-conditioned sparse linear systems~\cite{SGP:SGP06:061-070}. Each point in the oriented point cloud provides an estimate of the local gradient of the indicator function (which points are inside the surface of the object). A nice property of Poisson Surface solvers is that their solutions always produce watertight meshes, as they solve for an indicator volume, which produces a surface that is a $\mathbb{R}^2$ manifold folded in $\mathbb{R}^3$. This process can be done over a basis function set of B-splines with compact support.

\begin{figure}[bthp!]
\centering
\includegraphics[width=0.9\linewidth]{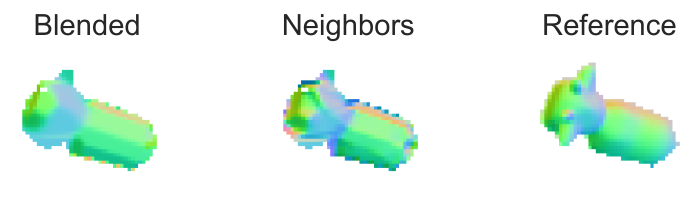} 
\label{fig:normals}
\vspace*{-3mm} 
\caption{\textbf{Visualization of Normals} for 3D Gaussians.}
\end{figure}

To produce an oriented point set, we simply go over all of our training views and render the point cloud of the object, to produce point locations $p = (x,y,z)$. The orientation of each point can be produced in two different ways (shown in \Cref{fig:normals}). The first, and most general, is to use the rasterization locality by taking a horizontal ($p_x$) and a vertical ($p_y$) screen-space neighbor for each point ($p$) and taking their cross product: 
\begin{equation}
    \vec{n} = \frac{(p-p_x) \times (p-p_y)}{|| (p-p_x) \times (p-p_y) ||} .
\end{equation}
This technique works for any differentiable renderer producing images on a grid (hence rasterization), but produces poor results on discontinuities. 

An alternative approach for 3D Gaussians is to re-use \Cref{eq:blend1} to blend all the local estimates of the normal ($\vec{n_i}$ instead of $t_i$) into a final estimate. The local estimate of the normal is given by the derivative of \Cref{eq:component}. 
\begin{equation}
    \vec{n_i} = \frac{\Sigma^{-1} (v t_i - \mu_i) }{|| \Sigma^{-1} (v t_i - \mu_i) ||} .
\end{equation}

In general, the sign and scale of the normal is fixed: the sign of the normal should face the camera created it, and the $|| \vec{n_i} || = || \vec{n_f} || = 1$. Both of these techniques can be seen in \Cref{fig:normals}. In practice, the blended definition is preferred as it requires no  neighbors. 

For producing meshes from our renderer, we typically perform the faster optimization with \Cref{sec:weightb}. and the rendering equation from \Cref{sec:alphac} can directly render that representation with no changes. The alpha compositing definition is preferred for mesh exporting as it is more view consistent than the heuristic blending. Lastly, we can reject any intersections that fail a quality threshold of a direct intersection, $\max_i(w_i) \leq \epsilon$ which is typically set to 0.9. The Poisson surface reconstruction produces a surface interpolation, and so the sparsity of point samples is not a problem. Additionally, we can export a colored, oriented point cloud, where the color is either the reconstructed color or the color of the images themselves at those points (see \Cref{fig:mesh_viz_fm}). We obtained object reconstructions by solving \Cref{eq:poisson} with Dirichlet boundary constraints~\cite{kazhdan2013screened}.

\begin{figure}[thp]
    \centering
    \begin{subfigure}[t]{\linewidth}        \centering

    \includegraphics[width=0.3\linewidth,trim={0 30 0 0},clip]{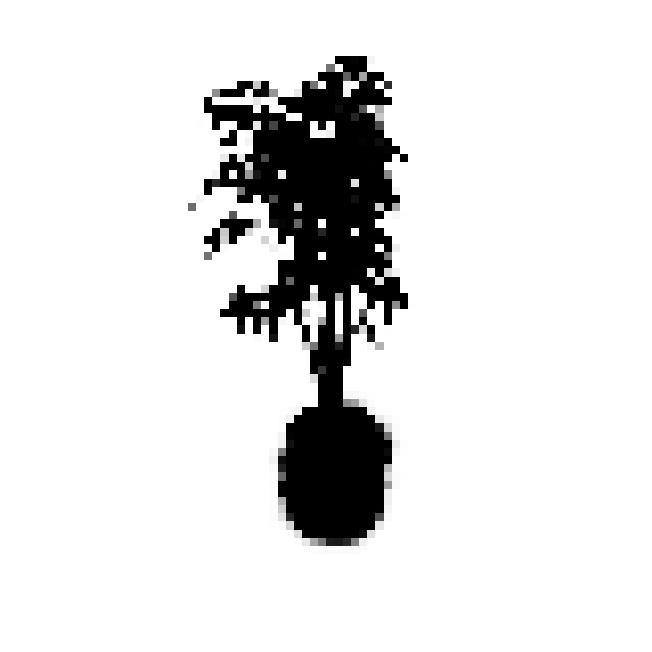}
    \includegraphics[width=0.3\linewidth,trim={0 30 0 0},clip]{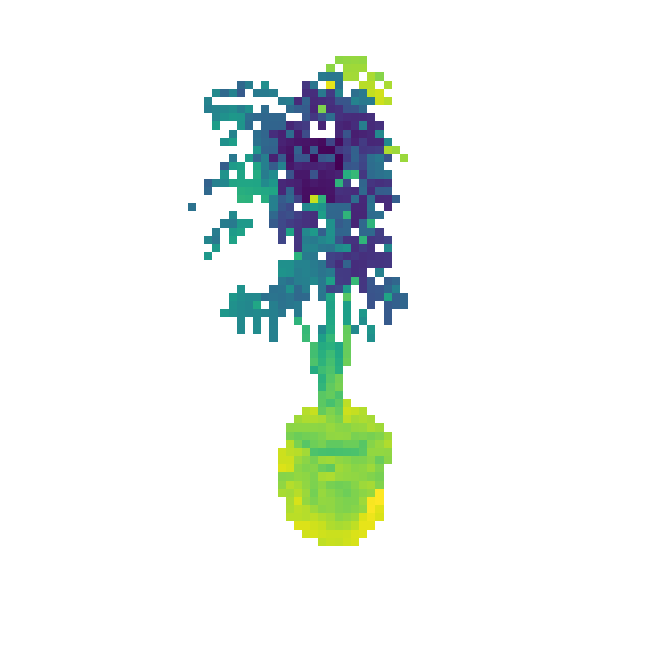} 
    \includegraphics[width=0.3\linewidth,trim={0 30 0 0},clip]{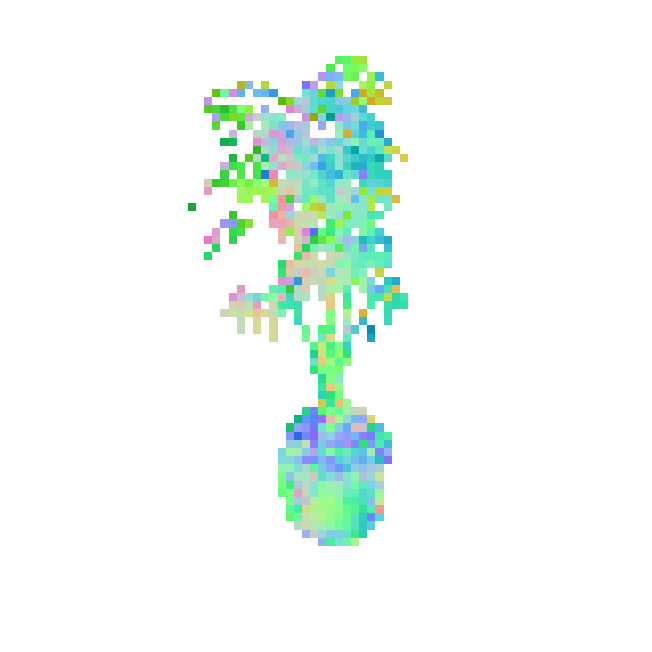}
    \caption{\textbf{Ficus via Weighted Blending (\Cref{sec:weightb})} }
    \end{subfigure}

    \begin{subfigure}[t]{\linewidth}        \centering

    \includegraphics[width=0.3\linewidth,trim={0 30 0 0},clip]{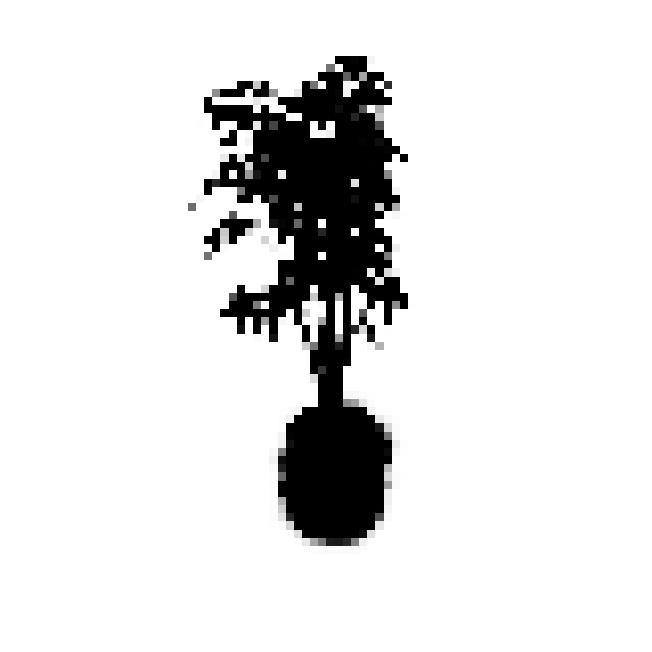}
    \includegraphics[width=0.3\linewidth,trim={0 30 0 0},clip]{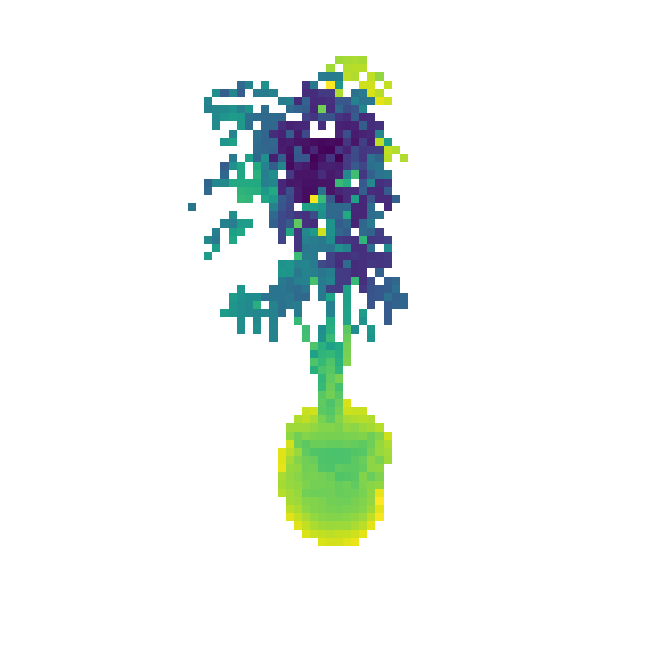} 
    \includegraphics[width=0.3\linewidth,trim={0 30 0 0},clip]{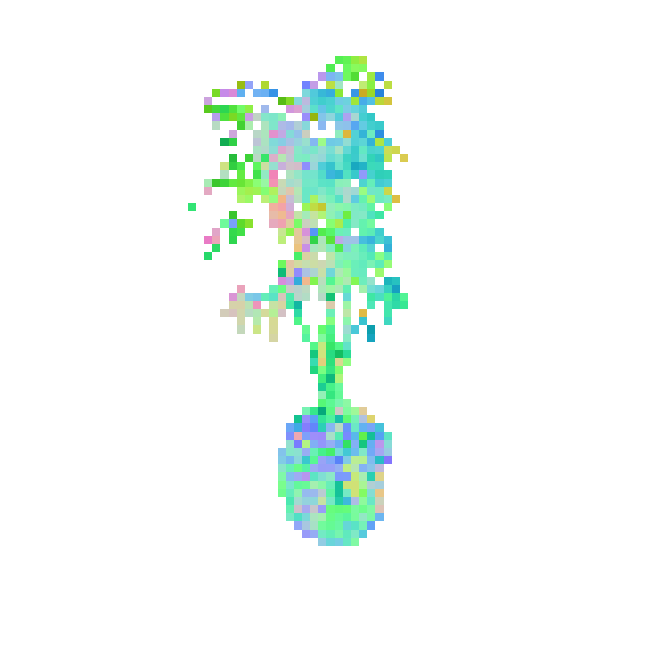} 
    \caption{\textbf{Ficus via Alpha Compositing (\Cref{sec:alphac})} }
    \end{subfigure}
\caption{\textbf{3D Gaussians are Fuzzy Metaballs} Reconstructed ficus from Gaussian Splatting~\cite{Kerbl2023splat}, rendered with Fuzzy Metaballs~\cite{keselman2022fuzzy}.  Shown are opacity, depth \& normals.} 
\label{fig:ficus_fm}
\end{figure}

\begin{figure}[tbhp]
    \centering
\begin{subfigure}[t]{\linewidth}
    \includegraphics[width=0.156\linewidth,trim={0 80 0 80},clip]{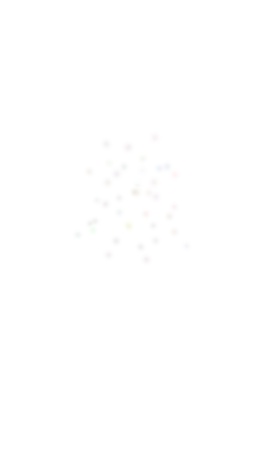}
    \includegraphics[width=0.156\linewidth,trim={0 80 0 80},clip]{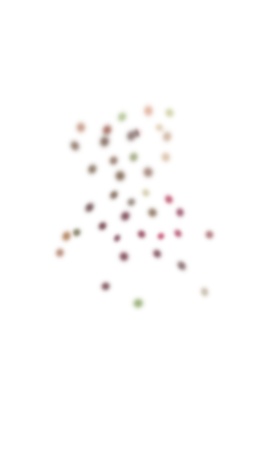}
    \includegraphics[width=0.156\linewidth,trim={0 80 0 80},clip]{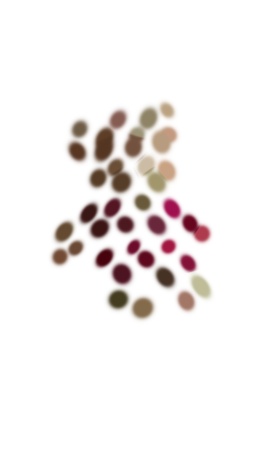}
    \includegraphics[width=0.156\linewidth,trim={0 80 0 80},clip]{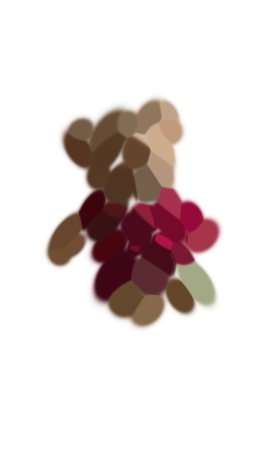}
    \includegraphics[width=0.156\linewidth,trim={0 80 0 80},clip]{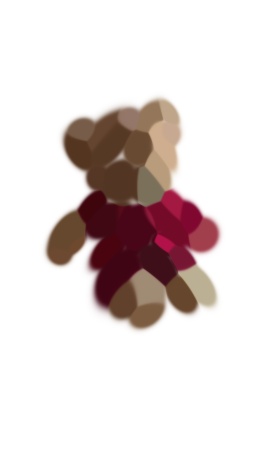}
    \includegraphics[width=0.156\linewidth,trim={0 80 0 80},clip]{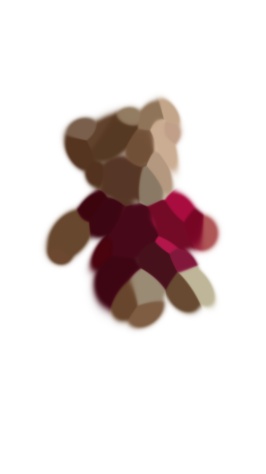}
    \includegraphics[width=0.156\linewidth,trim={0 80 0 80},clip]{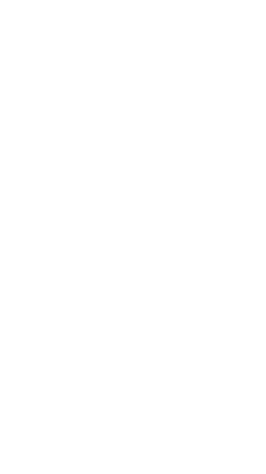}
    \includegraphics[width=0.156\linewidth,trim={0 80 0 80},clip]{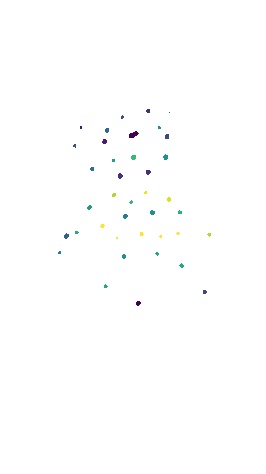}
    \includegraphics[width=0.156\linewidth,trim={0 80 0 80},clip]{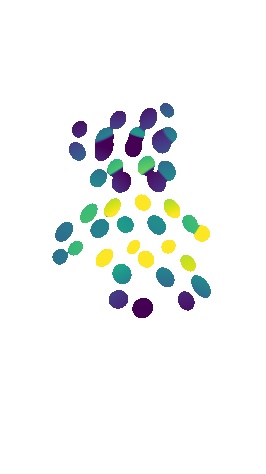}
    \includegraphics[width=0.156\linewidth,trim={0 80 0 80},clip]{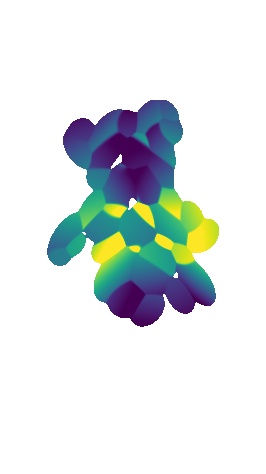}
    \includegraphics[width=0.156\linewidth,trim={0 80 0 80},clip]{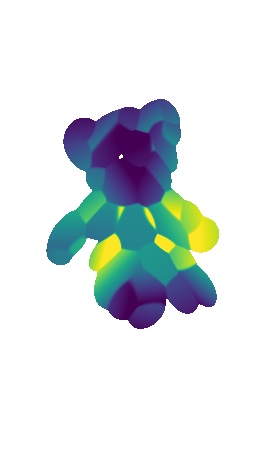}
    \includegraphics[width=0.156\linewidth,trim={0 80 0 80},clip]{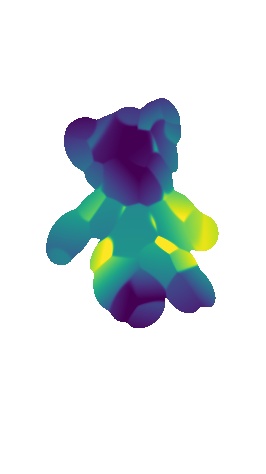}
    \caption{\textbf{Optimization via Weighted Blending (\Cref{sec:weightb})} }
    \label{fig:fm_opt2}
    \end{subfigure}
        \begin{subfigure}[t]{\linewidth}
    \includegraphics[width=0.156\linewidth,trim={0 80 0 80},clip]{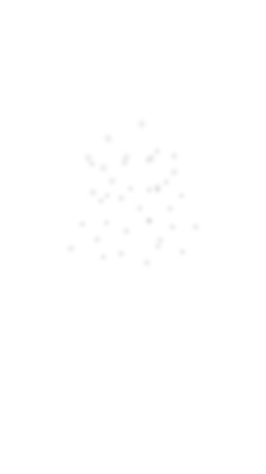}
    \includegraphics[width=0.156\linewidth,trim={0 80 0 80},clip]{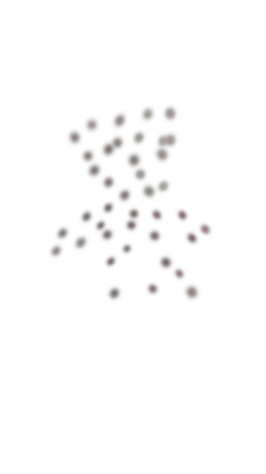}
    \includegraphics[width=0.156\linewidth,trim={0 80 0 80},clip]{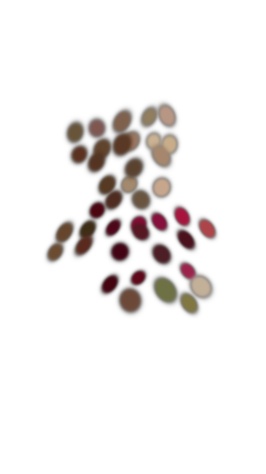}
    \includegraphics[width=0.156\linewidth,trim={0 80 0 80},clip]{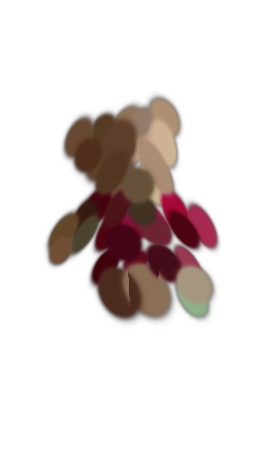}
    \includegraphics[width=0.156\linewidth,trim={0 80 0 80},clip]{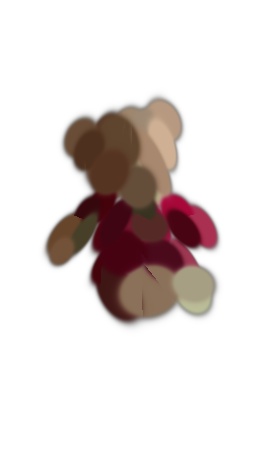}
    \includegraphics[width=0.156\linewidth,trim={0 80 0 80},clip]{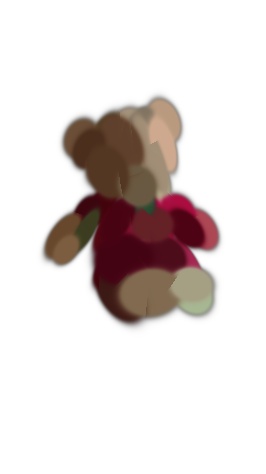}
    \includegraphics[width=0.156\linewidth,trim={0 80 0 80},clip]{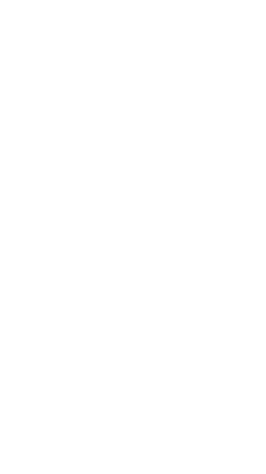}
    \includegraphics[width=0.156\linewidth,trim={0 80 0 80},clip]{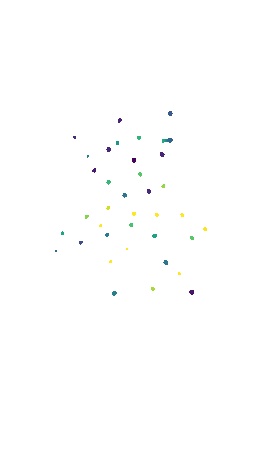}
    \includegraphics[width=0.156\linewidth,trim={0 80 0 80},clip]{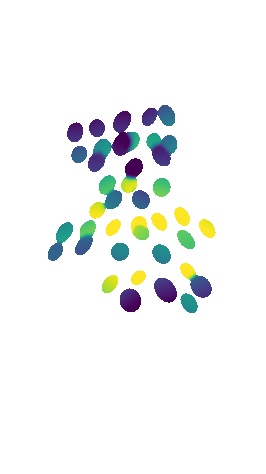}
    \includegraphics[width=0.156\linewidth,trim={0 80 0 80},clip]{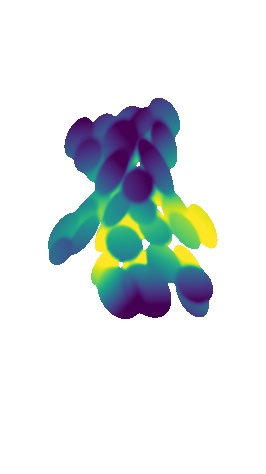}
    \includegraphics[width=0.156\linewidth,trim={0 80 0 80},clip]{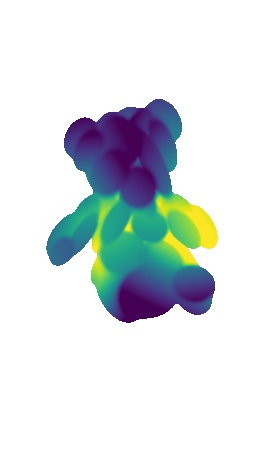}
    \includegraphics[width=0.156\linewidth,trim={0 80 0 80},clip]{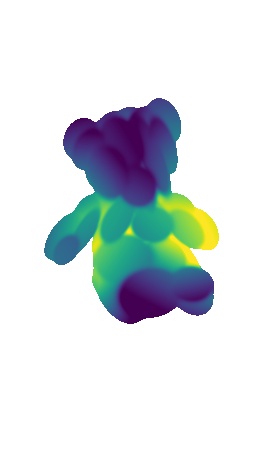}
    \caption{\textbf{Optimization via Alpha Compositing (\Cref{sec:alphac})} }
        \label{fig:fm_opt1}
    \end{subfigure}
\caption{\textbf{Both forms of rendering behave similarly.} Shown are 1\%, 4\%, 7\%, 12\%, 16\% and 20\% of the first epoch.}
\label{fig:fm_opt}
\end{figure}

\begin{figure*}[tbhp]
    \centering
\includegraphics[width=0.19\linewidth]{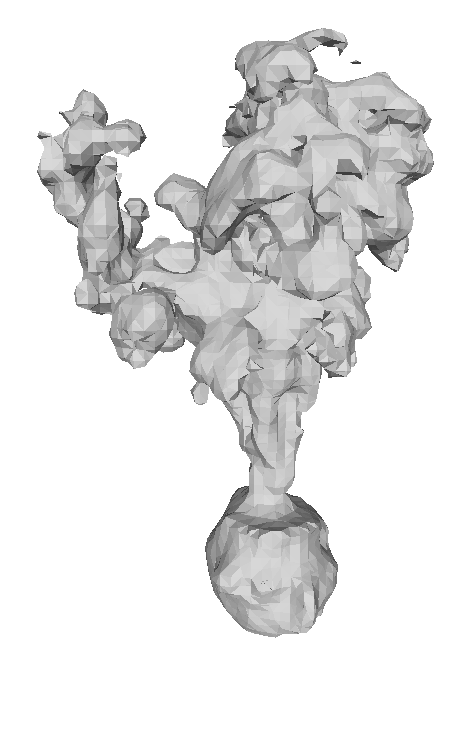} \includegraphics[width=0.19\linewidth]{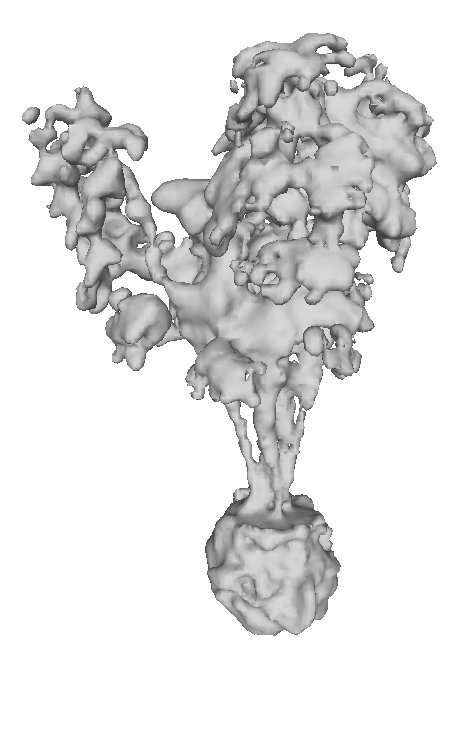} \includegraphics[width=0.19\linewidth]{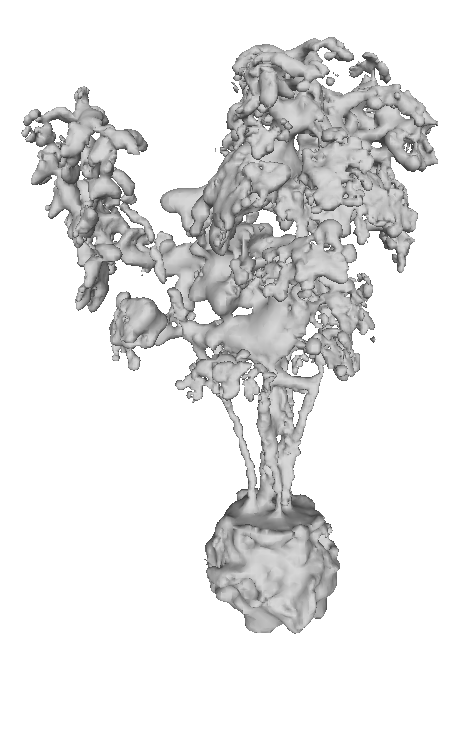} \includegraphics[width=0.19\linewidth]{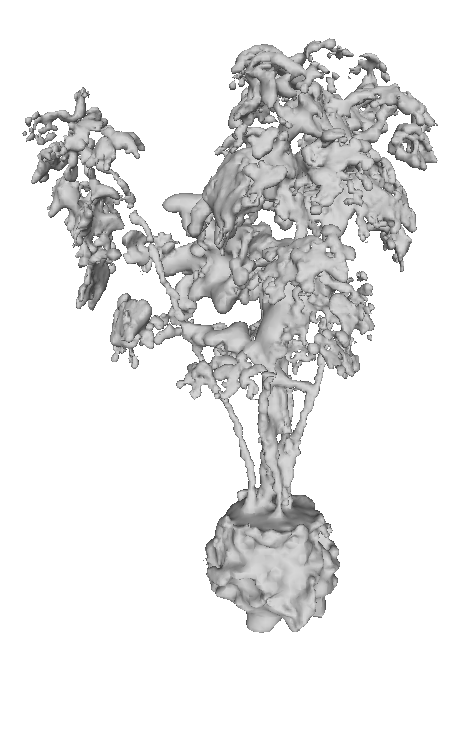} 
\includegraphics[width=0.19\linewidth]{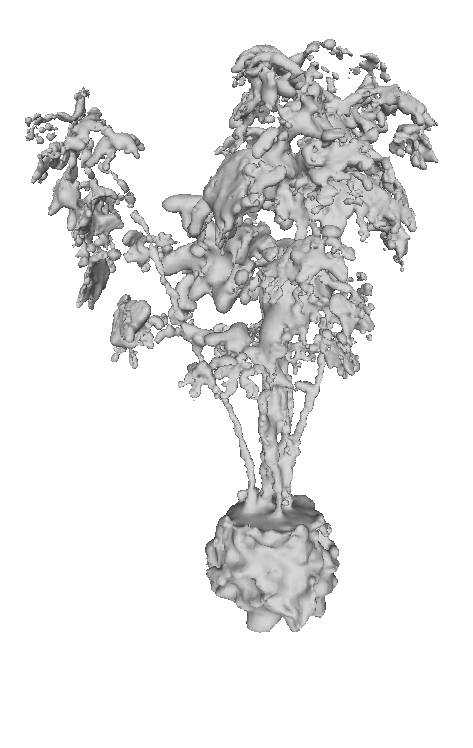}
\caption{\textbf{Visualization of Mesh Export} This ficus shape was reconstructed using the Gaussian Splatting~\cite{Kerbl2023splat} code but rendered as an oriented point cloud with Fuzzy Metaballs~\cite{keselman2022fuzzy} and reconstructed with a Poisson Solver~\cite{SGP:SGP06:061-070,kazhdan2013screened}. Left to right are Poisson tree depths of 6 through 10. While deeper tree depths produce more detailed reconstruction, noise and artifacts are also amplified. The mesh export is based on an oriented point cloud exported from 100 images of size $80 \times 60$. It is interesting to see that precise details like the various steps become visible, even when the forward passes were low resolution (see \Cref{fig:ficus_fm} for a visual example of the coarseness generated by the renderer).}
\label{fig:mesh_viz_ficus}
\end{figure*}

\section{Interoperability}\label{sec:interop}
We describe how Fuzzy Metaballs~\cite{keselman2022fuzzy} and 3D Gaussian Splatting~\cite{Kerbl2023splat} share a similar underlying shape representation. We demonstrate this by showing results in \Cref{fig:ficus_fm,fig:mesh_viz_ficus} where the initial shape reconstruction was performed using the 3D Gaussian Splatting paper~\cite{Kerbl2023splat}, and then directly converted and rendered using the methods in \Cref{sec:weightb,sec:alphac} and then exported using the oriented points using \Cref{sec:mesh}. 

Our experiments in \Cref{fig:fm_opt} and \Cref{sec:mesh} show that \Cref{sec:weightb,sec:alphac} use mutually compatible definitions of shape representation. However, the Gaussian Splatting~\cite{Kerbl2023splat} work uses an entirely different code base, in a different framework, optimizing scenes instead of objects, with no object masks, with a custom CUDA kernel, and preferring $\alpha$ to $\delta$ estimates. However, since both are using 3D Gaussians, we can render one with the other. 

We convert the Gaussian Splatting method to be compatible with our approach with only a few steps. Means are used directly, each $\Sigma$ are converted to $\Sigma^{-\frac{1}{2}}$, and the $\alpha$ for each Gaussian is replaced. For simplicity, we ignore $\alpha_i < 0.5$ and set $\lambda_i = \log(80)$ for the remaining 3D Gaussians, which we found works reasonably well\footnote{Inverse sigmoid conversion of $\lambda = -C \log(1-\alpha)$ maybe also be reasonable with appropriate $C$, which we did not search for. }. About 90\% of Gaussians had insufficient $\alpha$, creating a ten times speedup in our experiments.  For weighted blending experiments, we reused the settings of $\beta_1=21.4$ and $\beta_2=3.14$ from our prior experiments.

As can be seen in \Cref{fig:ficus_fm}, both techniques are able to capture the fine stem and leaf structures in the reconstructed ficus plant. The weighted blending technique demonstrates smoother normals, but the mesh export uses the alpha composting method and reasonable meshes can be obtained. 

In \Cref{fig:mesh_viz_ficus} we show Poisson surface reconstructions of the ficus, where the oriented point cloud used for solving the equation was generated by our renderer, but the original reconstruction was made with 3D Gaussian splatting. We show reconstructions at different tree depths, showing an increase in details, and an increase in noise, at finer scales. The solver can recover reasonably fine details considering considering the low resolution of the oriented point clouds. 

This extends the utility of the 3D Gaussian Splatting approach to be rendered with our fast methods that have viable JAX~\cite{jax2018github} CPU and GPU backends. Our approach enables  per-ray depth computation, normals and mesh exporting (\Cref{sec:mesh}). The ficus data was provided without color, so we show colored exports in \Cref{fig:mesh_viz_fm}. Mesh exports provides an interconnect with most 3D creation tools.  

\begin{table}[bth!]
\centering
\begin{tabular}{@{}lp{1.5cm}p{1.5cm}@{}}
\toprule
 & \textbf{CPU}  & \textbf{GPU}   \\ \midrule
Weighted Blending \\ (\Cref{sec:weightb})  & 4.94 $\mu$s & 226.6 ns \\
Alpha Compositing \\ (\Cref{sec:alphac}) & 12.7 $\mu$s & 377.6 ns \\ \bottomrule
\end{tabular}
\caption{\textbf{Runtime per ray, for an entire iteration} with a i5-7267U CPU and a GTX 1080 GPU. This was done with a 40 Gaussian model and these times include memory transfer times and forwards and backwards passes. Alpha compositing drops the need for hyper-parameters, in exchange for 200\% slower runtimes on CPU and 50\% slower runtimes on GPU. Since both approaches behave similarly (\cref{fig:fm_opt}) and are interoperable (\cref{sec:mesh}), this creates a trade-off between performance and simplicity. }
\label{tab:runtime_bck}
\end{table}

\begin{figure}[bthp!]
\centering
\includegraphics[width=0.9\linewidth]{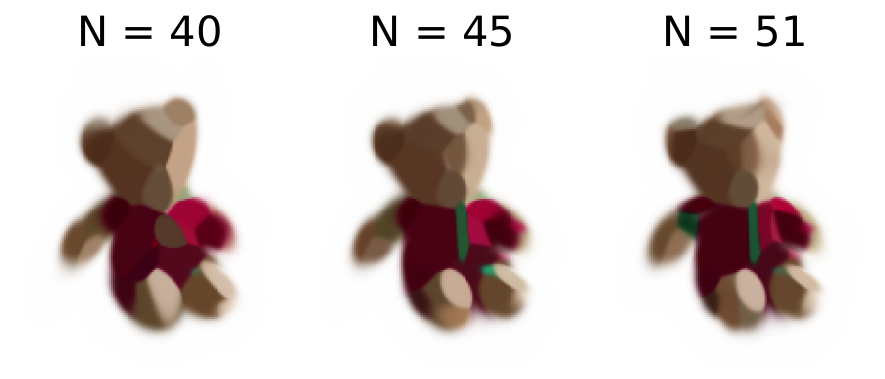} 
\caption{\textbf{Re-parameterized 3D Gaussians} to minimize apparent loss. $N$ is the number of mixtures used. The appearance of the tie and the sleeve in green can be seen with added Gaussians.  }
\label{fig:mixtures}
\end{figure}

\section{Splitting Gaussians}\label{sec:splits}
While the 3D Gaussian Splatting~\cite{Kerbl2023splat} explores some heuristic techniques for merging and splitting of Gaussians, here we develop an alternative, deterministic approach to modifying the number of Gaussians in the reconstruction.

We split 3D Gaussians after our initial model has converged according to statistical criteria~\cite{keselman2022fuzzy}. And we then repeat the optimization process. Two steps of this are shown in \Cref{fig:mixtures}. The splitting and removal process is based on removing Gaussians that contribute minimally to the reconstruction and splitting Gaussians that are given too much reconstruction responsibility based on the chosen loss.

To compute the set of Gaussians that minimally contribute, we compute the means and standard deviations of the weights assigned to each Gaussian $\mu_\lambda$ and $\sigma_\lambda$ and remove Gaussians that satisfy 
\begin{equation}
    \lambda_i \leq \mu_\lambda - z_\lambda \sigma_\lambda,
\end{equation}
where $z_\lambda$ is a z-score typically set to 2. 

To compute the splitting criteria, we take a random batch of rays (typically about 5\% of the dataset) and compute the per-ray loss (e.g. \cref{eq:loss} but any reconstruction loss is viable). Each ray also has an associated set of computed weights $w_i$ (\cref{eq:weight_alpha,eq:weight_new}). Over this batch of $M$ rays, we estimate the average loss associated with each Gaussian as $\bar{l_i} = \frac{1}{M} \sum^M_j = L_j \cdot w_i$. We compute the means and standard deviations of these losses as $\mu_{\bar{l}}$ and $\sigma_{\bar{l}}$ and split Gaussians that satisfy
\begin{equation}
    \bar{l_i} \geq \mu_{\bar{l}} + z_{\bar{l}} \sigma_{\bar{l}},
\end{equation}
where $z_{\bar{l}}$ is a z-score typically set to 1. 

Gaussians are split by forming two Gaussians by using the properties of the half-normal distribution~\cite{doi:10.1080/00401706.1959.10489866}. The two new means are generated by shifting the initial mean ($\mu_i$) in opposite directions by the direction of the dominant eigenvector ($\vec{v_1}$), and by the appropriate factor of the dominant eigenvalue ($\sigma_1$) of the covariance matrix ($\Sigma_i$):
\begin{equation}
    \mu_{a,b} = \mu \pm \sigma_1 \vec{v_1} \sqrt{\frac{2}{\pi}}.
\end{equation}
Both new Gaussians are given identical covariance matrices, reconstructed from the initial eigenvalues and eigenvectors but with a scaled dominant eigenvalue:
\begin{equation}
    \sigma_{1_{a,b}} =\sigma_1 \sqrt{ 1 - \frac{2}{\pi}}.
\end{equation}
The scaling factors $\sqrt{\frac{2}{\pi}} \approx 0.8$ and $\sqrt{1-\frac{2}{\pi}} \approx 0.6$ are based on the properties of the half-normal distribution. 

We replicate the initial weights ($\lambda$) and  colors ($c_i$) for the new Gaussians, with noise ($\epsilon_c = \epsilon_w = 0.1$) in their unconstrained parameterization space to avoid issues with coupled gradients during further optimization:
\begin{align}
    \log(\lambda_{a,b}) &= \log(\lambda) + \mathcal{N}(0,\epsilon_{w}),\\
    \sigma^{-1}(c_{a,b}) &= \sigma^{-1}(c) + \mathcal{N}(0,\epsilon_{c}).
\end{align}

This approach to splitting is deterministic and allows for increasing detail in an iterative way, as shown in \cref{fig:mixtures}. 

\section{Discussion}
With the increased deployment of vision systems in everyday environments, there is a need for flexible, efficient, and easily computed shape reconstructions. The recent developments in photorealistic differentiable rendering make close the dream of virtual systems accurately capturing everyday objects in the virtual world. 3D Gaussians, or metaballs, are a simple and powerful representation for shapes that enables easy reconstruction, as has been shown by prior work~\cite{Kerbl2023splat,keselman2022fuzzy,wang2023voge}. These techniques extend and interconnect these approaches and provide them additional flexibility. 

The equivalence of weighted blending and alpha composting approaches provides a wider array of options --- one being significantly faster and the other lacking hyper-parameters. The computation of per-ray blended normals allows for reliable mesh exporting, connecting to other 3D techniques and methods, without the need to pick thresholds for marching cubes~\cite{lorensen1987marching}, and producing watertight meshes via Poisson reconstruction~\cite{kazhdan2013screened}. We show a more deterministic, grounded approach to performing reparameterization of Gaussians.  Lastly, optical flow as a regularizer and prior for surface correspondences can easily and reliably improve the quality of shape reconstructions.

\section{Conclusion}
We have extended existing approaches for differentiable rendering of 3D Gaussians for speed, simplicity and flexibility. This expanded flexibility should allow these representations to be used in more places, for more applications, and potentially across a wider array of computational platforms than before. 

\clearpage
{\small
\noindent \textbf{Acknowledgements:} We would like to thank Jonathon Luiten for his seemingly boundless reservoir of curiosity, interest, and questioning; it proved essential prodding for developing parts of this work. Georgios Kopanas provided the 3DGS ficus and helpful conversations. Arkadeep Chaudhury, Jon Barron and Adam Harley provided additional beneficial conversations and feedback.
}

{\small
\bibliographystyle{ieee_fullname}
\bibliography{egbib}
}

\end{document}